\documentclass{article}
\pdfoutput=1
\usepackage[nonatbib,preprint]{nips_2018}

\usepackage[utf8]{inputenc} % allow utf-8 input
\usepackage[T1]{fontenc}    % use 8-bit T1 fonts
\usepackage{hyperref}       % hyperlinks
\usepackage{url}            % simple URL typesetting
\usepackage{booktabs}       % professional-quality tables
\usepackage{amsfonts}       % blackboard math symbols
\usepackage{nicefrac}       % compact symbols for 1/2, etc.
\usepackage{microtype}      % microtypography
\usepackage{subcaption}
\usepackage{afterpage}
\usepackage[flushleft]{threeparttable}
\usepackage{graphicx}
\graphicspath{{figures}}
\usepackage{amsmath}
\usepackage{amsfonts}
\usepackage{amssymb}

\usepackage[inline]{enumitem}

\usepackage{tikz}
\usetikzlibrary{arrows,shapes,arrows, arrows.meta, positioning,shapes.geometric}

\usepackage[eprint=false,maxbibnames=5,bibencoding=utf8,
backend=biber,style=ieee,citestyle=numeric-comp,
doi=false,isbn=false,url=false]{biblatex}

\bibliography{bcratos_ml_2023}

\usepackage[disable]{todonotes}

\title{Deep Learning for real-time neural decoding of grasp}

\author{
  Paolo Viviani \\
  LINKS Foundation\\
  Torino, Italy\\
  \texttt{paolo.viviani@linksfoundation.com} \\
\And
  Ilaria Gesmundo\\
  Politecnico di Torino\\
  Torino, Italy \\
  \And
  Elios Ghinato\\
  Politecnico di Torino\\
  Torino, Italy \\
\And
  Andres Agudelo-Toro\\
  DPZ, Deutsches Primatenzentrum GmbH\\
  Göttingen, Germany\\
  \texttt{aagudelo-toro@dpz.eu} \\
\And
  Chiara Vercellino\\
  LINKS Foundation\\
  Politecnico di Torino\\
  Torino, Italy\\
\And
  Giacomo Vitali\\
  LINKS Foundation\\
  Politecnico di Torino\\
  Torino, Italy\\
\And
  Letizia Bergamasco\\
  LINKS Foundation\\
  Politecnico di Torino\\
  Torino, Italy\\
\And
  Alberto Scionti\\
  LINKS Foundation\\
  Torino, Italy\\
\And
  Marco Ghislieri\\
  Politecnico di Torino\\
  Torino, Italy\\
\And
  Valentina Agostini\\
  Politecnico di Torino\\
  Torino, Italy\\
\And
  Olivier Terzo\\
  LINKS Foundation\\
  Torino, Italy\\
\And
  Hansjörg Scherberger\\
  DPZ, Deutsches Primatenzentrum GmbH\\
  Göttingen, Germany\\
}

\begin{document}

\maketitle

\begin{abstract}
  Neural decoding involves correlating signals acquired from the brain to variables in the physical world like limb movement or robot control in Brain Machine Interfaces. In this context, this work starts from a specific pre-existing dataset of neural recordings from monkey motor cortex and presents a Deep Learning-based approach to the decoding of neural signals for grasp type classification. Specifically, we propose here an approach that exploits LSTM networks to classify time series containing neural data (i.e., spike trains) into classes representing the object being grasped.\\
  The main goal of the presented approach is to improve over state-of-the-art decoding accuracy without relying on any prior neuroscience knowledge, and leveraging only the capability of deep learning models to extract correlations from data. The paper presents the results  achieved for the considered dataset and compares them with previous works on the same dataset, showing a significant improvement in classification accuracy, even if considering simulated real-time decoding.
\end{abstract}

% INTRO ==========================================================================
% ===================================================================================
\section{Introduction}\label{sec:intro}
    Neural decoding refers to the task of correlating signals recorded from the brain to variables in the outside world such as limb movement. This task is relevant for neuroscientists trying to understand the information contained in neural signal to improve our models of the brain, as well as for researchers developing Brain-Machine Interfaces (BMIs) to control physical and virtual objects (e.g., robotic prostheses, mouse cursors) \cite{pandarinathScienceEngineeringSensitized2021}. While the role of Machine Learning (ML) in neural decoding is known \cite{glaserMachineLearningNeural2020,livezeyDeepLearningApproaches2021}, improvements in this field can positively affect the quality of life of patients relying on BMIs to control their prosthesis (i.e., by providing a more natural control or by reducing the need of frequent re-training).

    This work was developed in the context of the B-Cratos EU project, which deals with
    % \footnote{\url{https://www.b-cratos.eu}}
    the real-time translation of intra-cranial brain signals, with the final goal of controlling a robotic prosthesis. In particular, this paper reports the results obtained in the simulated real-time decoding of the grasp type, based on a specific dataset previously recorded from the motor cortex of two Non-Human Primates (NHP) \cite{schaffelhoferDecodingWideRange2015}, investigated as a propaedeutic activity to the final prosthesis controller model deployment based on experimental data acquisition carried out within the project's scope. The main contributions proposed are:
    \begin{itemize}
        \item an LSTM model to detect the grasping phase from time series of neural data;
        \item an LSTM model to classify the object being grasped by the monkey that provides higher accuracy than the current state of the art for the same dataset.
    \end{itemize}
    Both models work on data provided as they would be in a real-time application: the grasping phase detection is representative of the capability to identify the beginning and the end of a grasp movement, while the classification of the object is used as a proxy for the grip type. Importantly, in the presented approach, the authors only focused on improving the predictive performance for practical applications without introducing any prior neuroscience knowledge in the model, and without the intent to provide any deeper understanding of neural activity and its correlation to the outside world.   

    The paper is structured as follows: section~\ref{sec:related} discusses the most relevant prior works that define the state of the art for neural decoding in general and this specific task in particular. Section~\ref{sec:dataset} and~\ref{sec:decoding} describe in detail the dataset used for this work and the ML approach for detecting movement and classifying the grasp type. Section~\ref{sec:results} reports and discusses the results obtained with the proposed approach, along with some considerations about the real-life application of these results. Finally, section~\ref{sec:conclusion} provides an overview about the relevance of these results and introduces ongoing research in the same direction.

% RELATED WORK ==========================================================================
% ===================================================================================
\section{Related work}\label{sec:related}
    Glaser et al. \cite{glaserMachineLearningNeural2020} and Livezey et al. \cite{livezeyDeepLearningApproaches2021} recently provided comprehensive reviews of the state of the art for ML and Deep Learning (DL) for neural decoding, observing how most of the applications still rely on very simple models (i.e., linear regression, Kalman filters) \cite{pandarinathScienceEngineeringSensitized2021,brandmanRapidCalibrationIntracortical2018}. These reviews also highlighted several works that have shown how LSTM can be effective to model the time evolution of neural signals \cite{yooLSTM,ahmadiDecodingHandKinematics2019}.
    Another interesting aspect is the different purposes that models for neural decoding can serve: if the goal is to understand the information contained in neural activity from a neuroscientist point of view, simple, explainable models are to be preferred over black-box machine learning algorithms, which are conversely more suited to provide strong predictive performance for BCI application. It should also be considered that advancements in explainable ML/DL may change this scenario.

    With respect to the decoding performance of ML models, since the purpose of this work is strictly related to the predictive performance on a specifc dataset, comparing this approach to prior works that dealt with completely different datasets is not trivial, nor particularly interesting for this work, as the task performed by the NHP, the recording conditions, and the pre-processing applied by each researcher can be completely different. As a consequence, the main reference benchmark for this work is a paper from Schaffelhofer et al. \cite{schaffelhoferDecodingWideRange2015}, which first tried to apply neural grasp decoding to this specific dataset. The approach was based on a naive Bayesian model used to classify the type of object being grasped by the NHP by looking at an entire region of interest in the neural recording (i.e., the phase when the NHP was holding the object). The accuracy obtained by Schaffelhofer et al. was of 62\%, on 50 classes (i.e., objects), evaluated using a leave-one-out cross validation technique.
    A recent thesis from F. Fabiani \cite{federicofabianiBraincomputerInterfaceBionic} discussed the results obtained by applying a sliding window to a longer region of the recordings, to simulate real-time operation. An LSTM network with a convolutional layer on top was used in this case, and the decoding results are reported not on the individual 50 objects, but grouping them by their shape (6 groups, variable size within groups) and size (6 groups variable shape within group): while the shape of the object was successfully classified with an accuracy of 92\% over 6 classes, the size decoding accuracy was around 25\% over 6 classes.
    Section~\ref{sec:results} will discuss how the presented approach differs from the previous works and how the results compare.

% DATASET ==========================================================================
% ===================================================================================
\section{Dataset structure}\label{sec:dataset}
    \subsection{Acquisition and original pre-processing}\label{sec:data_acquisition}
\begin{figure}[htp]
        \centering
        \includegraphics[width=0.7\textwidth]{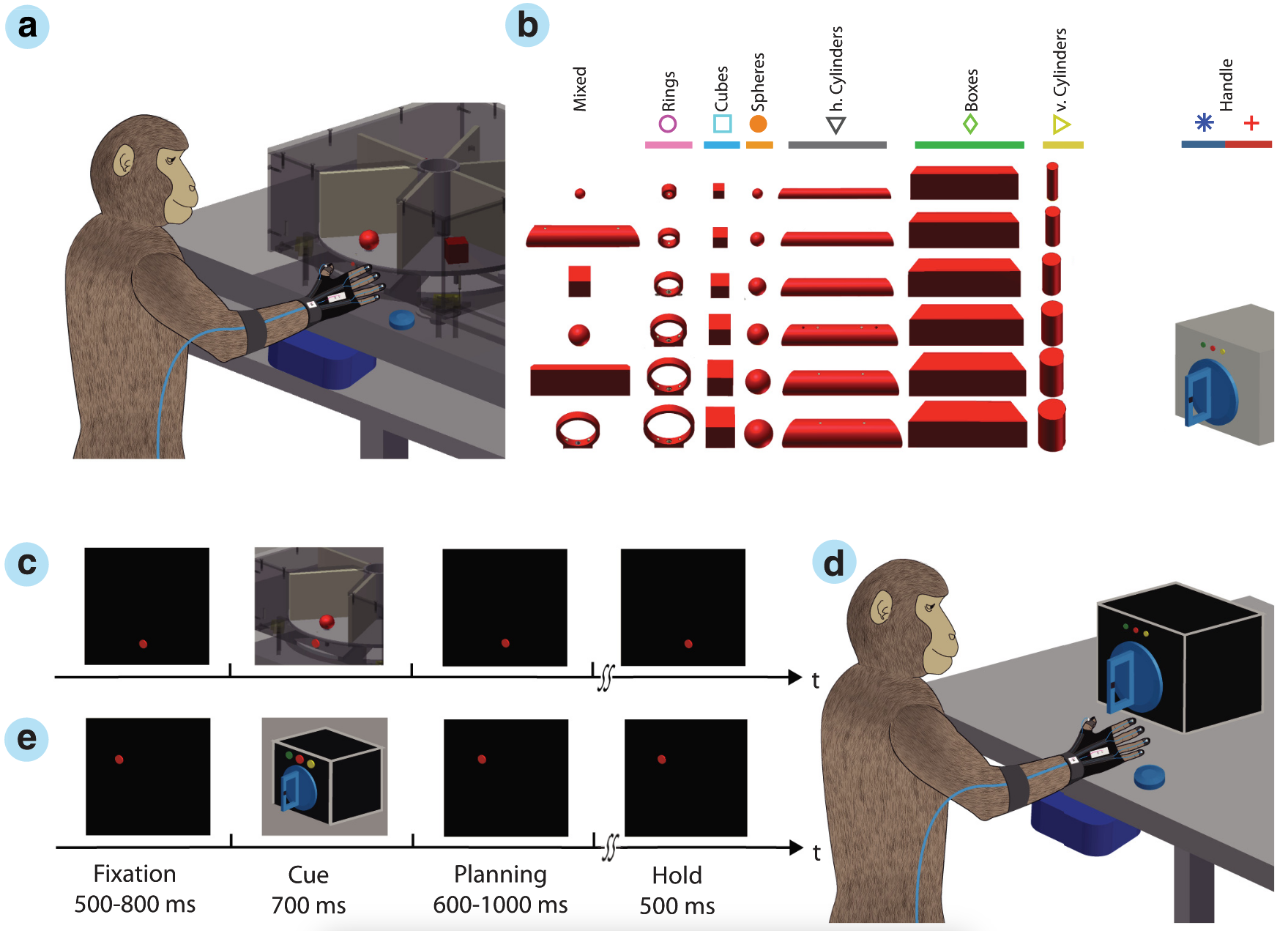}
        \caption{Experimental set up, adapted from \cite{schaffelhoferDecodingWideRange2015}. a) Two macaque monkeys were trained to grasp objects presented on computer-controlled turntable. b) The grasping objects are grouped in 50 classes. c) Objects were presented in pseudorandom order and the grasping taskfollowed pre-defined steps: Fixation; Cue; Planning; Movement and Hold. d,e) Adding two supplementary LEDs the monkey is prompted to perform a precision (yellow led) or a power (green led) grip on a special handle.}\label{fig:exp_setup}
    \end{figure}

    The dataset used for this work was acquired by researchers at Deutsches Primatenzentrum\footnote{\url{https://www.dpz.eu/en/home.html}} (DPZ) and involved recording neural signals from electrode arrays implanted in the AIP, F5, M1 regions of the brain cortex of two purpose-bred macaque monkeys (\emph{Macaca mulatta}), animal M and Z. The specific task performed by the NHPs is described in figure~\ref{fig:exp_setup} and involved grasping a set of objects at specific times, while recording the signal from 192 electrodes in the motor cortex.
    The original recording sampled at $24kHz$ has been processed through a band pass filter $(0.3 - 7 kHz)$ \cite{menzRepresentationContinuousHand2015}, then a \emph{spike sorting} algorithm has been applied offline to each recording session independently \cite{schaffelhoferDecodingWideRange2015}. Spike sorting algorithms recognise neurons activation patterns in the raw signal and can also identify multiple patterns to isolate multiple neurons from the same physical channel. The output of the spike sorting algorithm is defined as a multi-channel \emph{spike train}. A binary time series contains the information about activation time of each neuron (as identified by the spike sorting) during the experiment.
    For each animal involved in the experiment, three recording sessions acquired in during different days were made available, each one with a different number of spike channels (i.e., neurons) recorded due to the different outcome of the spike sorting algorithm for each session. The details about the datasets used are provided in section~\ref{sec:data_struct}.

    The process described in this section was performed out of the scope of the present work: only the final spike trains were available to the authors and are used here.

    \subsection{Structure}\label{sec:data_struct}
    \begin{figure}[htb]
        \centering
        \includegraphics[width=0.8\textwidth]{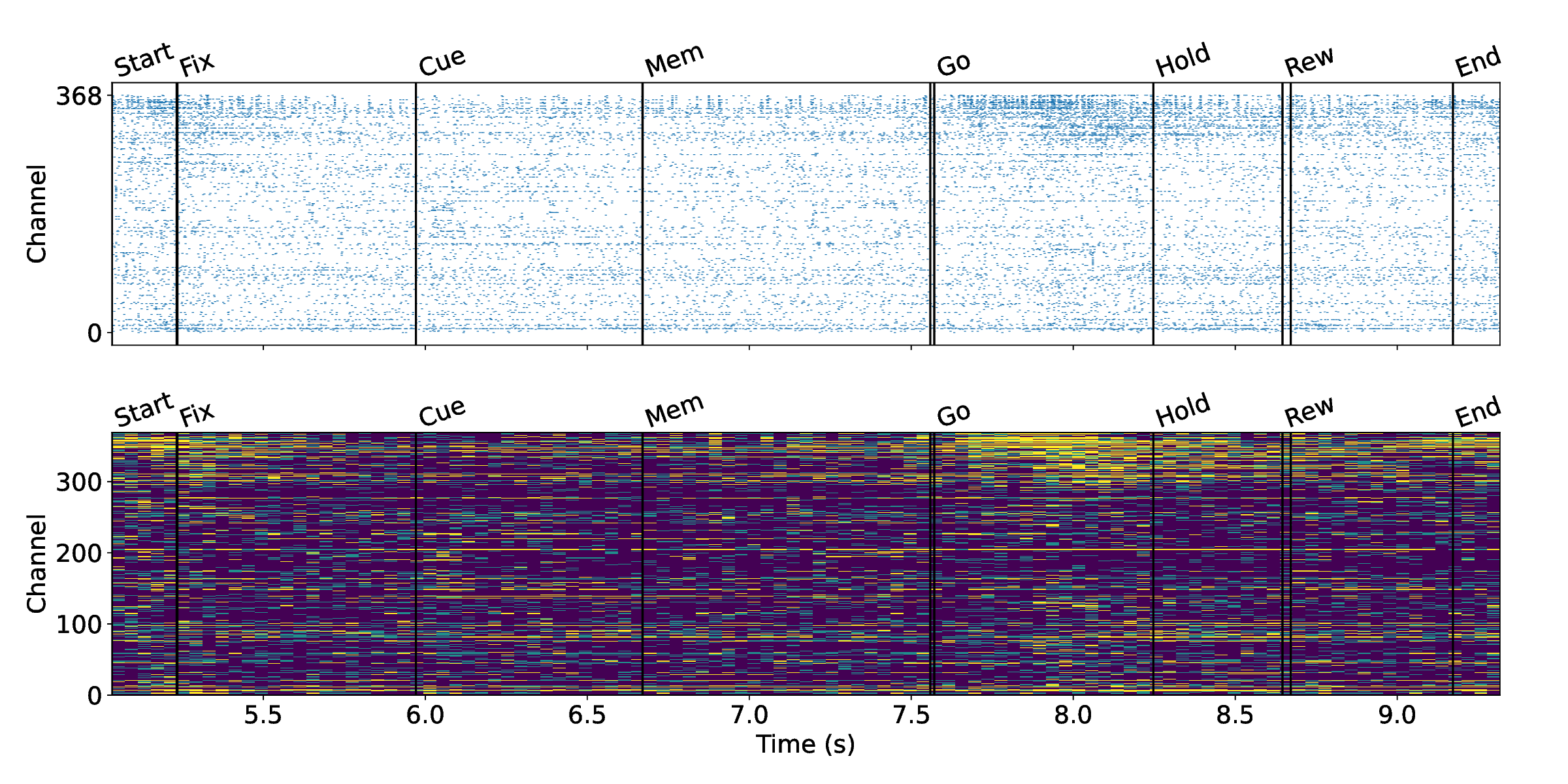}
        \caption{Example of recording for an experimental trial: the top plot reports the firing times for individual spikes for each channel, while the bottom plot reports the intensity map after a $40ms$ time discretisation. The color of each bin is correlated to the number of spikes counted in that time interval and in that channel (increasing from purple to yellow). Time on $x$ axis is reported from the beginning of the recording session.}\label{fig:spike_train}
    \end{figure}
    \begin{table}[htb]
        \begin{center}
        \caption{Number of channels and number of trials for each experimental recording session.}\label{tab:datasets}
        \begin{tabular}{|c|c|c|c|}
        \toprule
        NHP identifier &  Dataset identifier & \# Channels & \# Trials \\
        \midrule
         & MRec40 & 552 & 745\\
        M & MRec41 & 568 & 757\\
         & MRec42 & 554 & 653\\
        \midrule
         & ZRec32 & 391 & 687\\
        Z & ZRec35 & 388 & 724\\
         & ZRec50 & 369 & 610\\
        \bottomrule
        \end{tabular}
        \end{center}
    \end{table}
    The datasets corresponding to the six recording sessions were provided as Neo \cite{neo14} objects, with multi-channel spike trains segmented in \emph{trials}, each one corresponding to an instance of the grasp/release process described in figure~\ref{fig:exp_setup}. Figure~\ref{fig:spike_train} represents the recording of the spike trains for a single trial from animal Z, plotting both the spike trains (top) and the same data discretised in $40ms$ time bins (bottom), where each bin contains the number of spikes counted in that time interval for that channel. As also shown in figure~\ref{fig:spike_train}, timestamps for each experiment phase, corresponding to subsequent stages of the monkey task (e.g., visual cue, planning, grasping, etc.) are reported in the data, as well as the information about the object being grasped (\emph{object id}). It should be noted that not all the phases are reported in figure~\ref{fig:exp_setup} and that time intervals between phases can change between trials due to subject behaviour. 

    Table~\ref{tab:datasets} reports the number of channels identified by the spike sorting for each recording session, along with the dataset identifier and the number of trials for that session.
    It should be also noted that the number of times each object appears in the dataset is not constant among objects and recording sessions, resulting in a class imbalance when training a classifier based on grasped objects.
    
% APPROACH ==========================================================================
% ===================================================================================
\section{Decoding approach}\label{sec:decoding}
\subsection{Pre-processing}\label{sec:preprocessing}
    The proposed approach is aimed at simulating on-line decoding of grasped object (used as a proxy for the grasp type), hence the data should be prepared to fit this purpose. The pre-processing steps are split as follows, to separate the different steps and support the development of models without the need of re-processing the whole dataset every time, as intermediate data structures are always stored. These steps are performed separately for each recording session of each animal.
    \paragraph*{1) Time discretisation} The spike trains are discretised in bins of $40ms$ of length. Also shorter time binning has been evaluated to understand if longer, more detailed sequences could benefit the classification performance, but due the lack of evidence of a positive impact, shorter sequences and larger time bins have been selected to lower the computational cost.
    Each trial was converted to a dense matrix with an ID and stored to disk, and the related metadata was stored in a dedicated data structure (progressive bin number associated to experiment phases; object id; data matrix id). 
    \paragraph*{2) Training, validation and test split}
    The split of each dataset in training, validation and test sets is critical: in fact, since the sequences used to train the model are partially overlapping, splitting data at sequence level would lead to an artificially high accuracy as highly overlapped sequences can appear in all sets. To avoid this effect and provide realistic validation of the approach, the split was performed at the entire trial level (i.e., entire trials were assigned to either train, validation or test set), setting apart 80\% of the trials for training, 16\% for validation, and 20\% for testing. Moreover, class-based stratification\footnote{\url{https://scikit-learn.org/stable/modules/cross\_validation.html\#stratification}} was performed to ensure sufficient representation of all classes in all the datasets.

    In this phase, under-represented classes (i.e., objects appearing in less than 3 trials for each recording session) were removed from the dataset to ensure the presence of at least one representative of the class in each dataset partition. Moreover, object of identical shape and size are present in multiple groups (c.f. columns of figure~\ref{fig:exp_setup}b) and were mapped to the same class.
    \paragraph*{3) Sequence creation}
    After splitting the trials in the three datasets, a \emph{sliding window} was applied to each trial. The sub-sequences extracted by the sliding window represent the data a real-time decoder would see at any given moment, and were tagged either with a label indicating that the monkey was in rest position, or with the grasped object id if the last bin of the sequence was falling in the \emph{region of interest} that has been defined as the \emph{hold} phase of the experiment. This choice was made to be coherent with the approach from Schaffelhofer et al.
    
\begin{figure}[htb]
        \centering
        \includegraphics[width=0.6\textwidth]{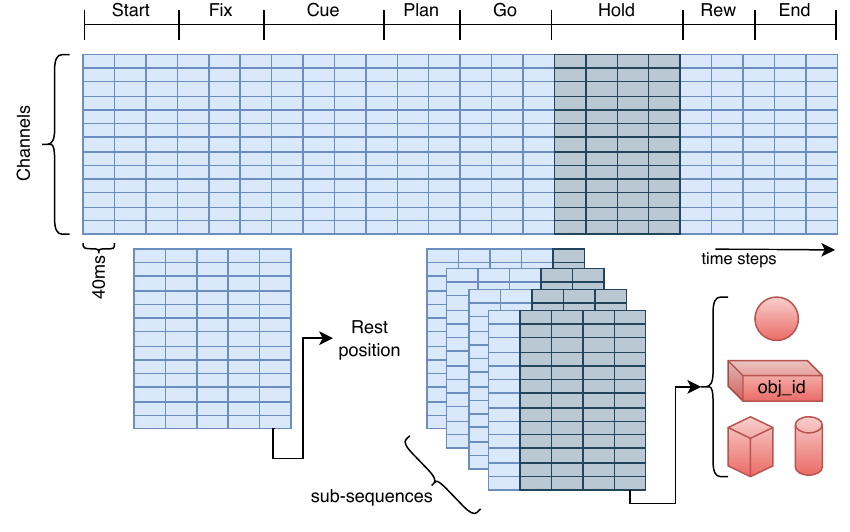}
        \caption{Illustration of the sequence creation process. Each discretised trial is scanned with a sliding window, advancing one time step at a time and extracting fixed-length sequences (2D arrays of dimension $channels\times length$). Each sequence is stored along with a label that can be \emph{rest}, if the last bin falls outside of the \emph{hold} phase, or the object id if it falls within. }\label{fig:preproc}
    \end{figure}

    The extraction of sequences is depicted in figure~\ref{fig:preproc}. From the sub-sequences, two different datasets are created that will be used for different learning tasks:
    \begin{itemize}
        \item an \emph{grasping phase} dataset, that contains all the sequences labelled as \emph{grasp} (if they fall in the region of interest) and \emph{rest} (elsewhere). Class imbalance is roughly 10 to 1 for rest vs. grasp;
        \item a \emph{classification} dataset, containing only the sequences with the last bin falling in the region of interest, labelled with the id of the grasped object.
    \end{itemize}
    These two datasets (for each recording session) are then passed on to the model building step, discussed in the next section.
\subsection{Decoding Model}
    Among the two tasks identified, the classification one has  proven to be significantly more challenging than the grasping phase one. All the effort to identify a suitable architecture was then focused on the classification task, and the best performing model was re-used for the grasping phase detection task.
    \subsubsection{Architecture}
    Based  on previous literature, it has been decided to start the search for a suitable architecture from LSTM networks. Additional experiments were performed within the scope of this work with Bidirectional LSTM networks \cite{schusterBidirectional,bidirectionalLSTM}, that demonstrated significantly better performance than simple LSTM networks.

\begin{figure}[htb]
        \centering
        \includegraphics[width=0.45\textwidth]{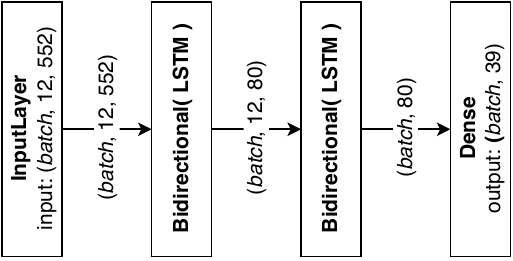}
        \caption{Final architecture for classification network (animal M). Activation functions for the LSTM layers are $\tanh$ and sigmoid (recurrent activation). Output layer activation is softmax. Loss function is categorical crossentropy.}\label{fig:architecture}
    \end{figure}

    These experiments included a preliminary architecture and hyperparameter search that led to identify the following subset of hyperparameters, which have been used to run a final optimisation campaign for each dataset with KerasTuner \cite{omalley2019kerastuner} bayesian optimiser; the search space was defined as in table~\ref{tab:hyperparams}.
    An example of model architecture is shown in figure~\ref{fig:architecture}.
    \begin{table}[htb]
        \centering
        \caption{Final hyperparameter search space. One model identified for each animal.}\label{tab:hyperparams}
        \begin{tabular}{|l|l|c|c|}
        \toprule
        Hyperparameter & Values & \multicolumn{2}{c|}{Selected} \\
        & \footnotesize{(L1=0.01, L2=0.01)} & M & Z\\
        \midrule
        LSTM layers & \{ 1, 2, 3, 4 \} & 2 & 1\\
        Hidden units & \{ 16, 32, 40, 64 \} & 40 & 40 \\
        Dropout & \{ 0, 0.2, 0.4, 0.6, 0.7, 0.8 \} & 0.8 & 0.7\\
        Kernel regularisation & \{ None, L1, L2, L1 + L2 \} & L2 & L2\\
        Recurrent regularization & \{ None, L1, L2, L1 + L2 \} & L2 & L1+L2\\
        Initial learning rate & \{ $10^{-3}$, $2\cdot10^{-4}$, $10^{-4}$ \} & $10^{-3}$ & $10^{-3}$ \\
        \bottomrule
        \end{tabular}
    \end{table}
    \subsubsection{Training}
    The model was implemented with Keras \cite{chollet2015keras}, to facilitate its future deployment with Tensrflow lite on a dedicated embedded board\footnote{\url{https://coral.ai/docs/dev-board-mini/datasheet/}}. Training used a mini-batch size of 256 and typically converged within 100 epochs, which are almost never reached due to an early stopping policy based on the evolution of validation accuracy. A learning rate (LR) scaling policy was implemented to halve the LR when loss is plateauing for more than 10 epochs. 
    % RESULTS ==========================================================================
% ===================================================================================
\section{Results and discussion}\label{sec:results}
    \subsection{Grasping phase detection}\label{sec:res_activation}
    This is a binary classification task with a significant class imbalance towards the \emph{rest} class. The LSTM model reaches an accuracy of at least 98\% for all datasets, the F1 score is always greater than 0.95. Table~\ref{tab:activation} presents both the confusion matrix for one dataset (MRec40) and the accuracy figures for all datasets: if considering this model in the context of a finite-state prosthesics control scenario, unwanted movements (rest classified as grasp) would amount to 1\% of the time steps considered, while unresponsiveness (grasp classified as rest) is around the 0.1\% of time steps. The relevance of these results for a real-life prosthesics control is discussed in section~\ref{sec:conclusion}.
    \begin{table}[htb]
        \begin{center}            
        \caption{Grasping phase detection results.}\label{tab:activation}
        \begin{subtable}{0.38\textwidth}
            \centering
            \caption{Confusion matrix for grasping phase detection (MRec40).}\label{tab:activation_cm}
            \begin{tabular}{|c|cc|}
            \toprule
            & \multicolumn{2}{c|}{Predicted} \\
            True & rest & grasp \\
            \midrule
            rest & 12824 & 173 \\
            grasp & 25 & 1347 \\
            \bottomrule
            \end{tabular}
        \end{subtable}
        \begin{subtable}{.58\textwidth}
            \centering
            \caption{Grasping phase detection accuracy metrics.}\label{tab:activation_acc}
            \begin{tabular}{|l|c|c|}
            \toprule
            Dataset id & Accuracy & F1 score \\
            \midrule
            MRec40 & 99\% & 0.96 \\
            MRec41 & 99\% & 0.96 \\
            MRec42 & 99\% & 0.97 \\
            ZRec32 & 99\% & 0.96 \\
            ZRec35 & 98\% & 0.96 \\
            ZRec50 & 98\% & 0.95 \\
            \bottomrule
            \end{tabular}
        \end{subtable}
        \end{center}
    \end{table}

    \subsection{Grasped object classification}\label{sec:res_classification}
\begin{figure}[htb]
     \centering
     \begin{subfigure}{\textwidth}
         \centering
         \includegraphics[width=0.85\textwidth]{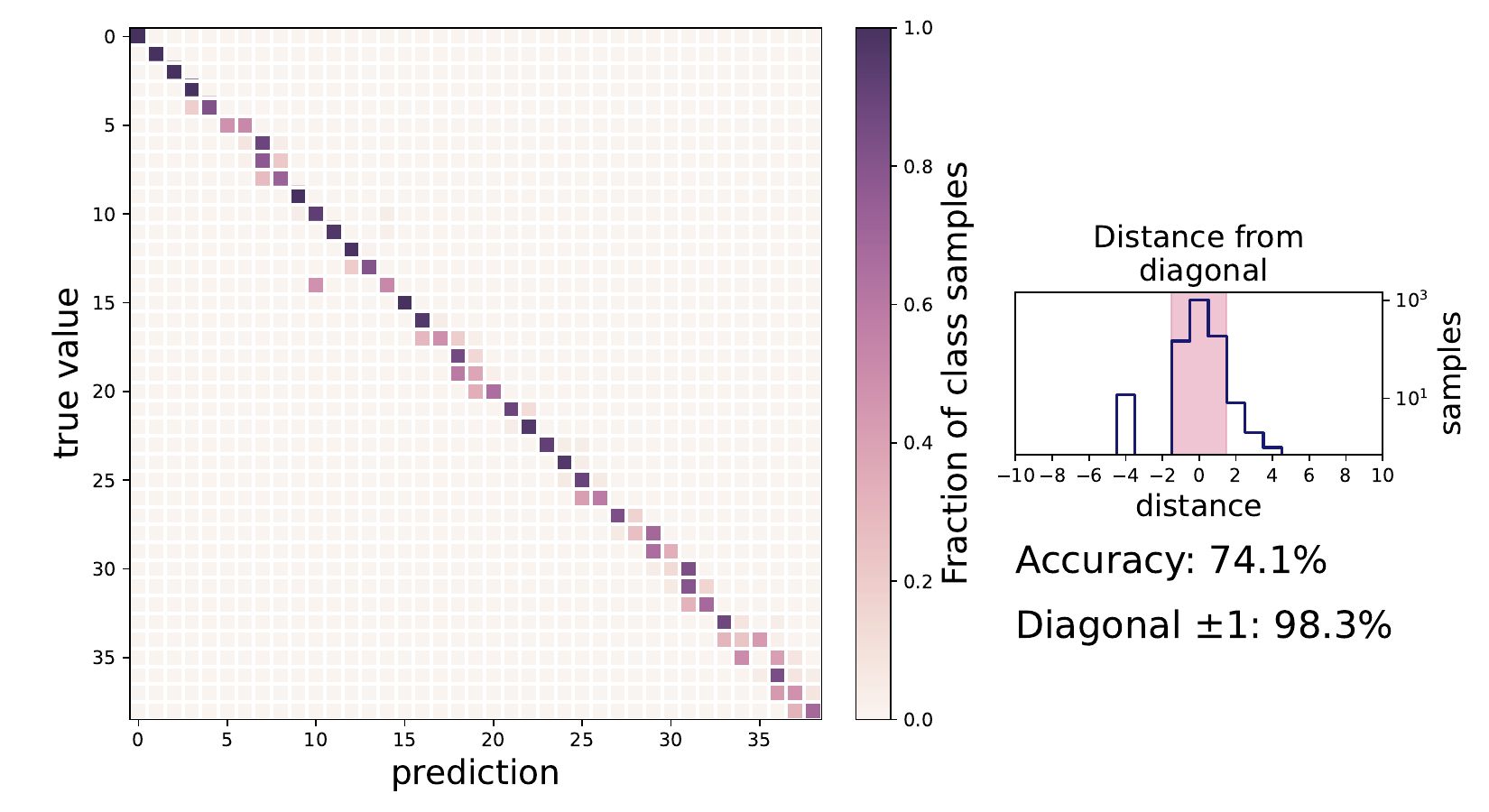}
         \caption{Results for MRec40}
     \end{subfigure}
     \begin{subfigure}{\textwidth}
         \centering
         \includegraphics[width=0.85\textwidth]{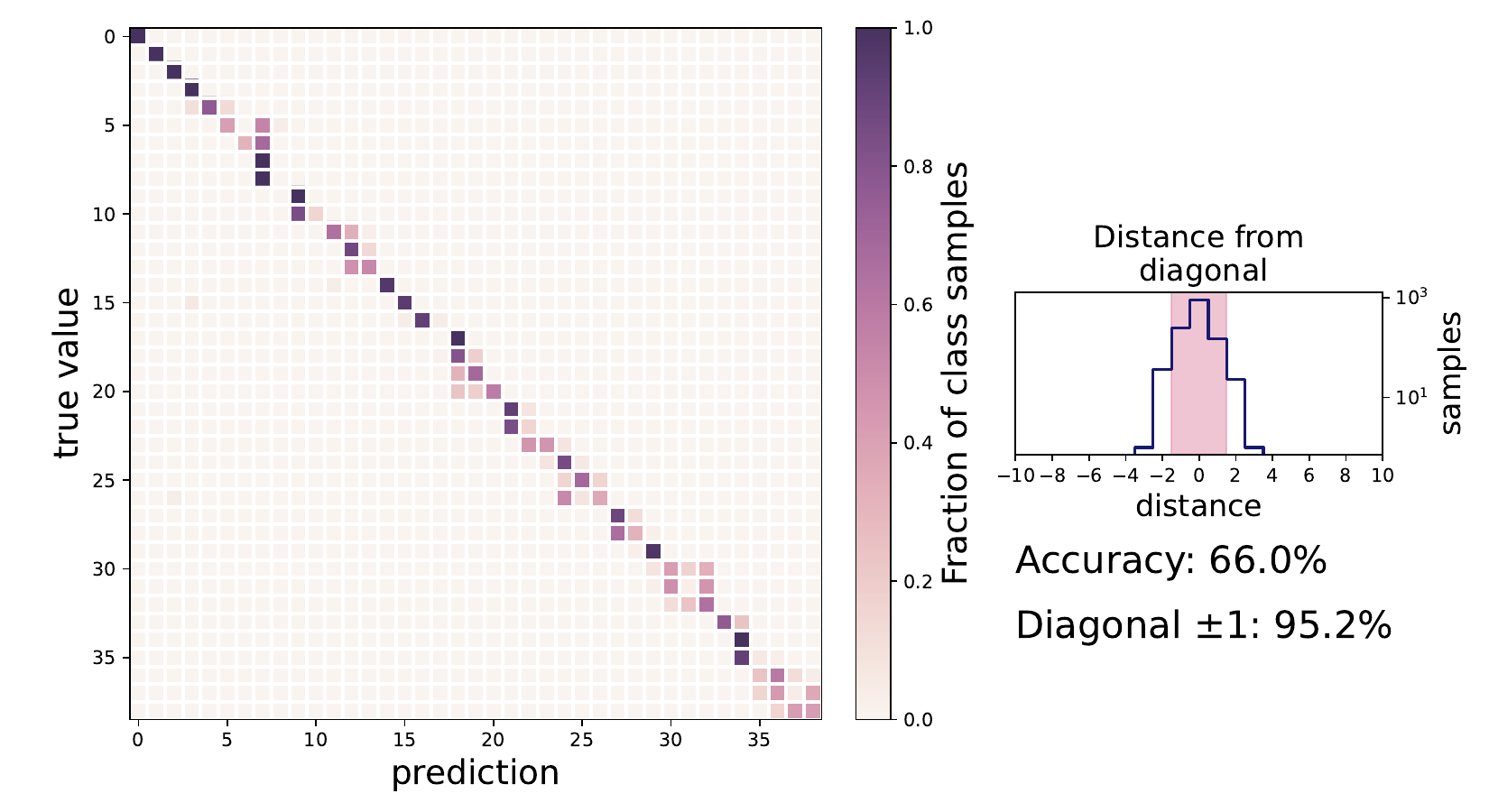}
         \caption{Results for MRec41}
     \end{subfigure}
     \begin{subfigure}{\textwidth}
         \centering
         \includegraphics[width=0.85\textwidth]{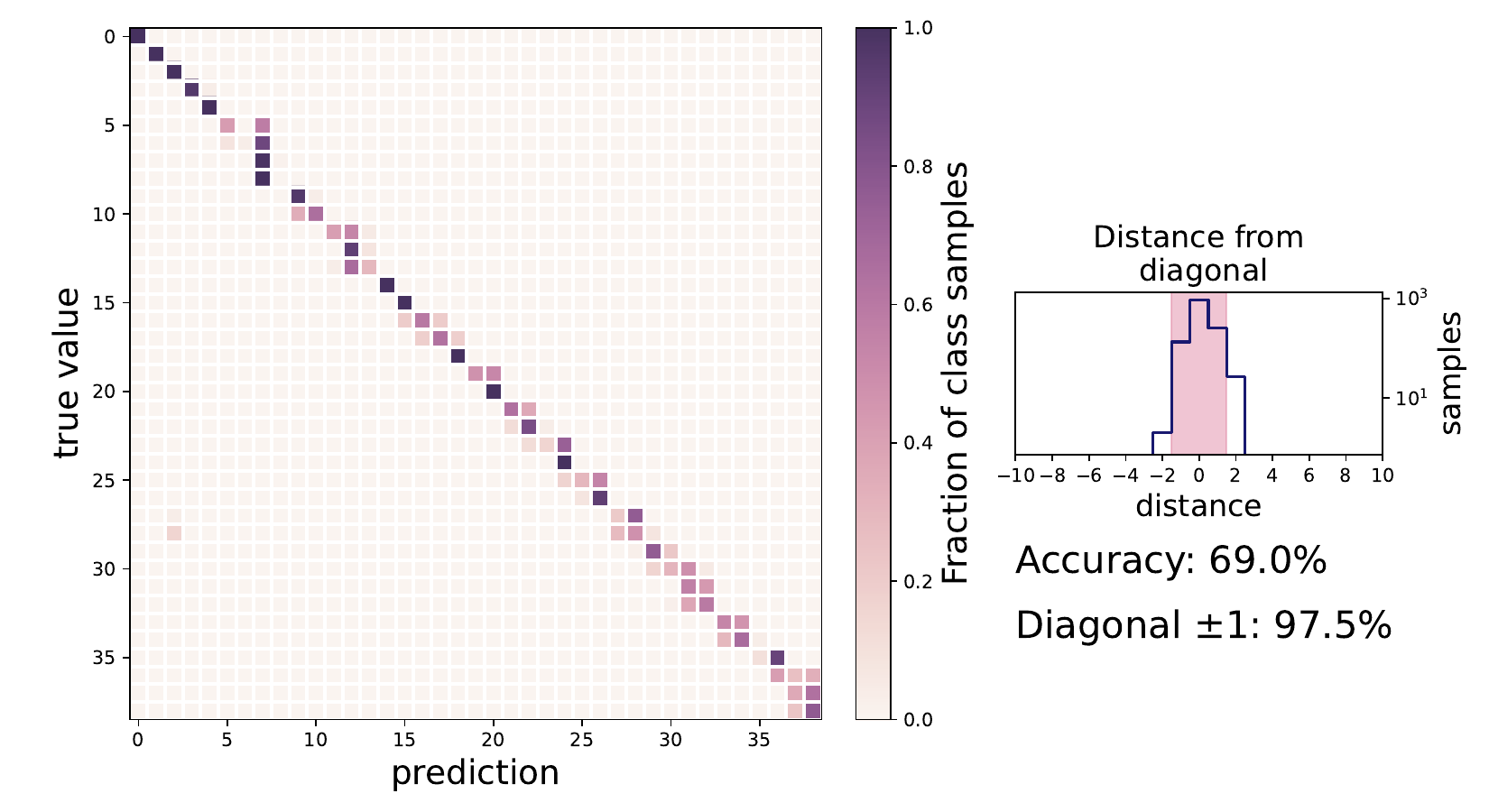}
         \caption{Results for MRec42}
     \end{subfigure}
        \caption{Confusion matrices and accuracy results for grasped object classification for animal M. The models used are the best performing on the validation set among 10 training runs. It can be noted how most of the misclassified samples lie very close to the diagonal.}
        \label{fig:m_results}
    \end{figure}

\begin{figure}[htb!]
        \centering
        \begin{subfigure}{\textwidth}
            \centering
            \includegraphics[width=0.85\textwidth]{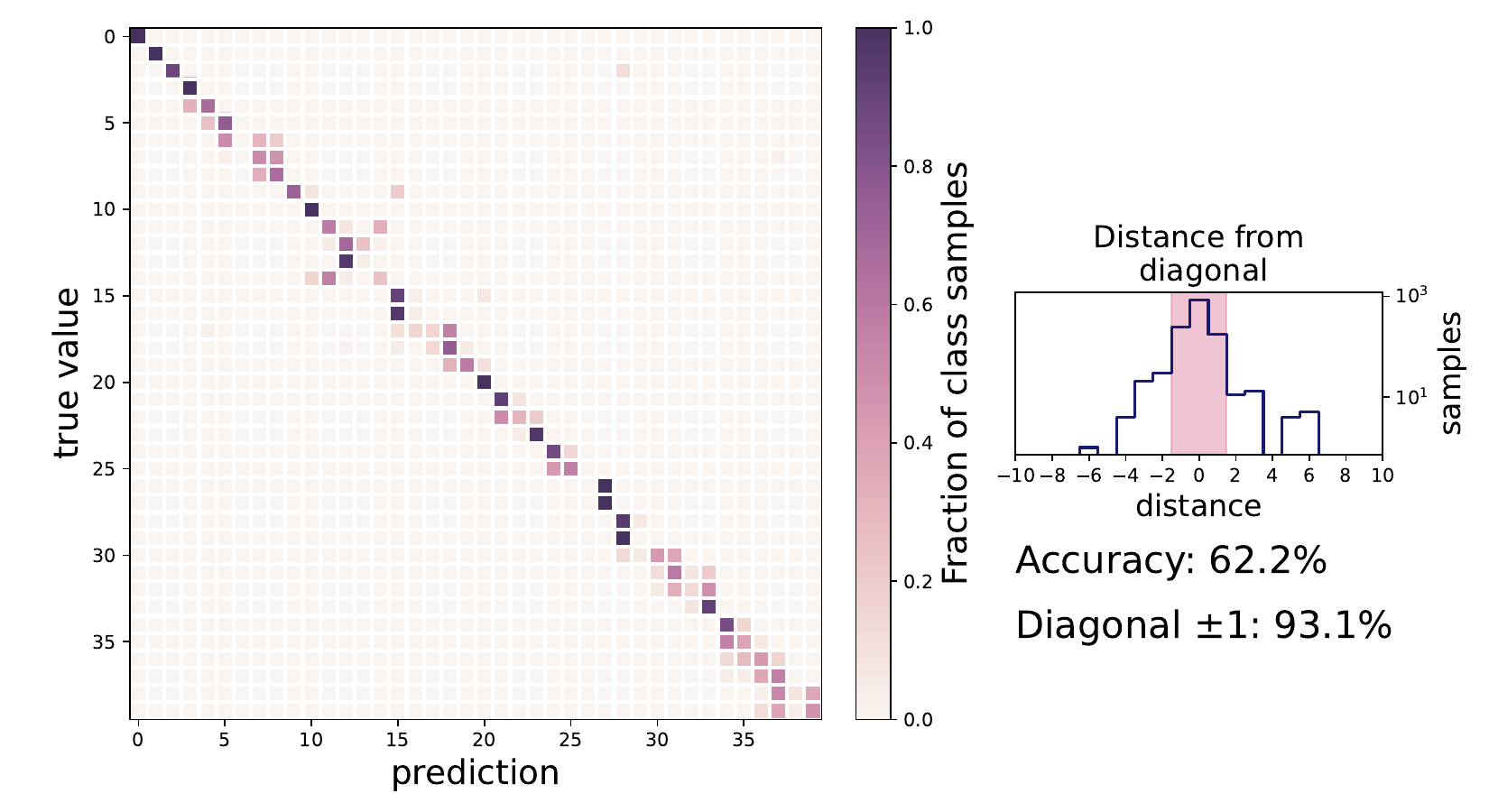}
            \caption{Results for ZRec32}
        \end{subfigure}
        \begin{subfigure}{\textwidth}
            \centering
            \includegraphics[width=0.85\textwidth]{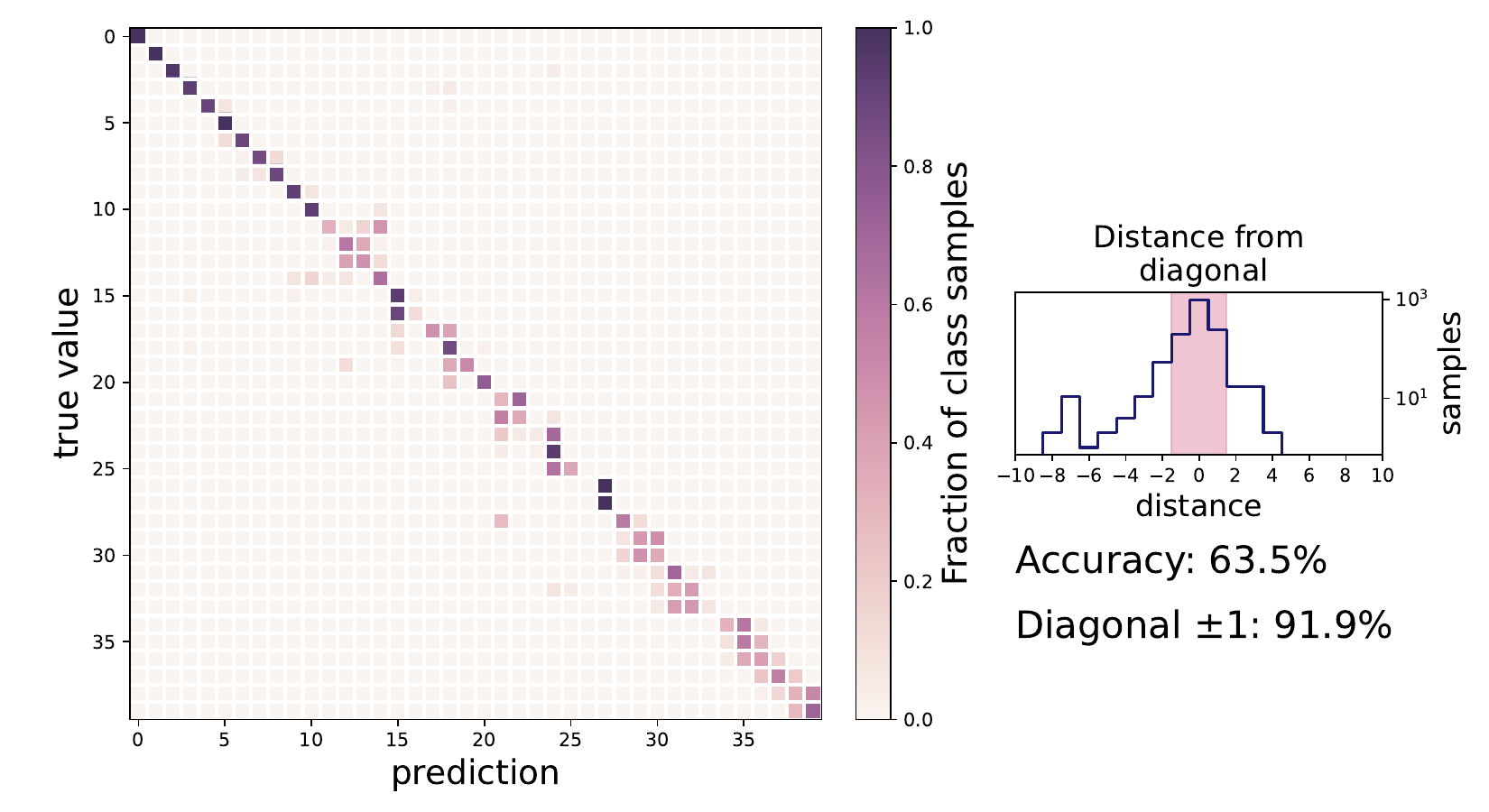}
            \caption{Results for ZRec35}
        \end{subfigure}
        \begin{subfigure}{\textwidth}
            \centering
            \includegraphics[width=0.85\textwidth]{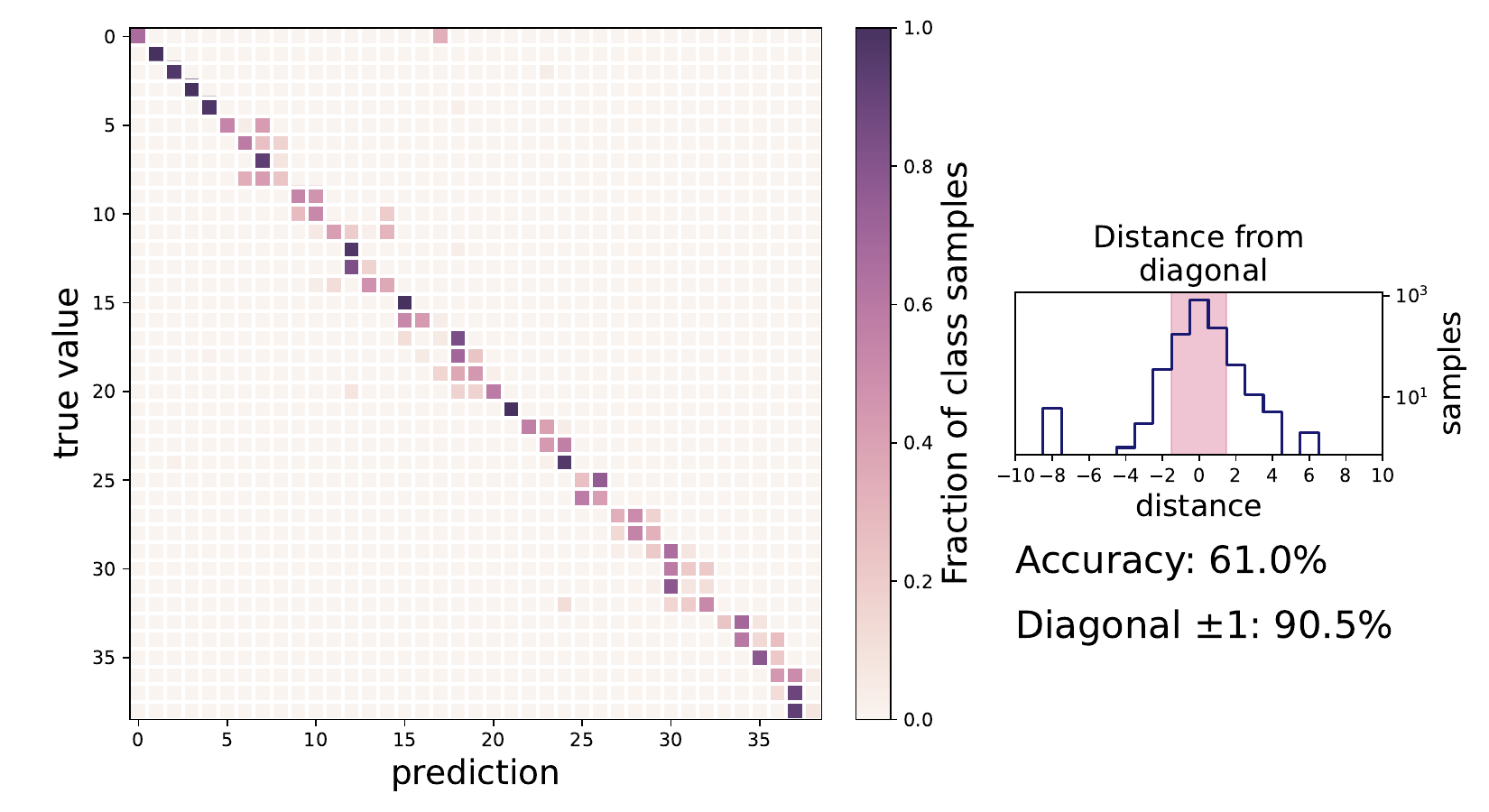}
            \caption{Results for ZRec50}
        \end{subfigure}
           \caption{Confusion matrices and accuracy results for grasped object classification for animal Z. The models used are the best performing on the validation set among 10 training runs. It can be noted how most of the misclassified samples lie very close to the diagonal.}
           \label{fig:z_results}
       \end{figure}

    The results for the classification of the grasped object represent the major contribution of this work and figure~\ref{fig:m_results} and~\ref{fig:z_results} report the confusion matrices and accuracy figures for each recording session available to the authors. 
    To make a sensible comparison with prior art, we should carefully consider the differences between the presented approach and the previous works mentioned in section~\ref{sec:related}. 
    
    In the original 2015 paper \cite{schaffelhoferDecodingWideRange2015}, Schaffelhofer et al. report an average accuracy for the \emph{hold} phase of $62\%$ over a total of 10 recording sessions (against the six available for this work). This was obtained with an off-line naive bayesian classifier applied to the whole \emph{hold} phase and validated with a leave-one-out (LOO) approach, hence the dataset fraction used for training was significantly higher than what is used in this work. On the other hand, 50 classes were considered while in this case under-represented classes (i.e., less than 3 samples) were removed, and identical objects with different IDs were collapsed onto the same class. In general it is possible to say that the presented approach outperforms the naive bayes classifier (significantly for animal M, and slightly for animal Z) in a harder set-up (sliding window vs. fixed region; smaller training partition). Additionally, as classes are ordered as the objects in figure~\ref{fig:exp_setup}, neighboring classes represent the same object with a slightly different size, hence a very similar grip type; in this sense it has been computed also the \emph{relaxed accuracy}, counting 1-class-away misclassification as correct. Relaxed accuracy results are also reported in table~\ref{tab:res_comp} and figure~\ref{fig:m_results},~\ref{fig:z_results}, showing a better performance than previous works. 

    With respect to the results reported by Fabiani \cite{federicofabianiBraincomputerInterfaceBionic} for the online decoding, the approach is very similar (i.e., sliding window), although a larger region of interest was considered, including also the \emph{go} phase. In this case results are not reported by individual objects, but by shapes and sizes (respectively rows and columns in figure~\ref{fig:exp_setup}b): to compare the accuracy with the numbers presented in this paper, shape and size accuracy figures are multiplied. The results obtained here are even outperforming the offline decoding algorithm also proposed by Fabiani, which considers a fixed, larger window of each trial for the classification.
    The comparison of table~\ref{tab:res_comp} shows that also in this case, the presented LSTM model outperforms significantly the previous approach, which can be surprising given that LSTM models are used in both cases; it is the author's opinion that the adoption of Bidirectional LSTM layers provide the most significant improvement in the decoding performance, along with careful usage of regularisation strategies (i.e., a high dropout fraction proved to be particularly effective).  

       \begin{table}[htb]
        \centering
        \caption{Comparison of results with previous works. Metrics are averaged over the available recording session for the present work (standard deviation is referred to multiple recording sessions, not to multiple training shots) and for Schaffelhofer et al \cite{schaffelhoferDecodingWideRange2015}. The work by F. Fabiani only reports results grouped by object type and object size, the individual object accuracy is calculated as the product between the two accuracies \cite{federicofabianiBraincomputerInterfaceBionic}. Relaxed accuracy also considers as correctly classified objects belonging to the first neighboring classes.}\label{tab:res_comp}
        \begin{tabular}{|l|l|c|c|c|}
        \toprule
        Animal & Metric & Present work & Schaffelhofer, 2015 & Fabiani, 2021 \\
        \midrule
        M & Accuracy & $69.7 \pm 4\%$ & $62.9 \pm 3.6\%$ & n/a \\
        \midrule
        Z & Accuracy & $62.3 \pm 1.2\%$ & $61.4\pm4.1\%$ & n/a \\
        \midrule
        Global & Accuracy & $65.9 \pm 4.9\%$ & $62\%$ & 22\% \\
        & Relaxed accuracy & $94.4 \pm 3.1\%$ & $86.5\%$ & 59\% \\
        \bottomrule
        \end{tabular}
    \end{table}

    \subsection{Real-life usage and current deployment}
    \subsubsection{Reduced training set}
    When testing real-time neural decoders in the context of real NHP experiments, one of the requirements is the capability to quickly start decoding signals and shortnening the training data acquisition as much as possible. Moreover, neural signal evolves through time (i.e., due to mechanical movement of the implant, inflammatory processes, brain plasticity, etc.), making re-training models daily mostly unavoidable.

    Here are reported the results of an investigation aimed at understanding how the size of the training set w.r.t. the entire dataset is affecting the accuracy of the trained model. This is particularly relevant as gathering samples is a time consuming operation in both neuroscience research with non-human primates and BMI clinical practice with patients.
    In this sense, the model capability to retain a good accuracy is critical to ensure its usability in the real world. Table~\ref{tab:decrease_train} reports the accuracy numbers for an increasing test set partition (and an equally decreasing training+validation set). Qualitatively the accuracy remains quite close to the previous state of the art up to a 30\% training partition; moreover, the relaxed accuracy is also better than the state of the art down to 30\% of training data.
    \begin{table}[htb]
    \begin{center}
    \begin{threeparttable}
        \caption{Accuracy and relaxed accuracy for a progressively reduced training set.}\label{tab:decrease_train}
        \begin{tabular}{|c|c|c|}
        \toprule
        Training + validation set & Accuracy & Relaxed accuracy \\
        \midrule
        80\% & 74.1\% & 98.3\% \\
        70\% & 70.1\% & 97.1\% \\
        60\% & 69\% & 93.7\% \\
        50\% & 62.8\% & 94\% \\
        40\% & 63.8\% & 92.8\% \\
        30\%$^*$ & 59.2\% & 87.5\% \\
        20\%$^{**}$ & 51\% & 81\% \\
        \bottomrule
        \end{tabular}
        \begin{tablenotes}
        \item\scriptsize{$^*$validation is 30\% of training set to ensure at least one representative per class}
        \item\scriptsize{$^{**}$validation is 40\% of training set} 
        \end{tablenotes}    
    \end{threeparttable}
    \end{center}
    \end{table}
    \subsubsection{Computational cost}
    The training time for all the grasp classification cases is lower than one minute on an Nvidia V100, suitable for on-the-fly re-training when new data becomes available during experiments. The larger number of sequences used to train the grasping phase detection model makes each epoch longer, but the number of epochs required to reach a satisfactory accuracy is smaller and contributes to keep the training time at around one minute.

    \subsubsection{Current deployment of the model}
    This work represents a preparatory step toward an effective model for real-time continuous control of a prosthesis: while this is a classification task, the data recorded from the brain is fully representative of the kind of data that will be acquired within the scope of the project during 2023, and the network architecture that has been identified will be the starting point for the deployment of a real-time decoder on a dedicated device. Since the final model will predict a limited number of real-valued degrees of freedom, it has been built a synthetic dataset reproducing a trajectory of a single degree of freedom based on the recordings used in this work and the same model presented in figure~\ref{fig:architecture} has been trained to predict this real value with good accuracy. It is in fact currently being used to control the opening and closing of a robotic hand for the first integration tests.

    \subsubsection{Reproducibility note}
    The code used to achieve the presented results is available in a public repository\footnote{\url{https://github.com/LINKS-Foundation-CPE/Neural-decoding-paper-2023}}. Instructions to access a subset of the data (one recording session) has been made available to reviewers as additional material through the submission portal, and such dataset can be tested directly with the code just mentioned. The entire dataset has been provided by DPZ out of courtesy as part of an ongoing collaboration in the B-Cratos project, but the ownership remains to the original researchers and its complete publication is beyond the scope of this work.
\afterpage{\clearpage}

% CONCLUSION ==========================================================================
% ===================================================================================
\section{Conclusion and future work}\label{sec:conclusion}
    The experimental results presented in the previous section demonstrated the capability of bidirectional LSTM networks to match and outperform state-of-the-art methodologies on known data. While this dataset, being used for a categorical classification task, is not immediately relevant for the continuous control of a robotic prosthesis, it validates the suitability of LSTM models for neural decoding of hand grasp actions. In particular, the very good results obtained for the relaxed accuracy are more representative of the performance to be expected when controlling a limited number of degrees of freedom or less nuanced categories (i.e., 5/6 grasp types, opening and closing). It is also worth to highlight how these results were obtained without involving any prior neuroscience knowledge.

    With respect to the next steps, the main research activity ongoing is directed towards making LSTM models robust to the evolution of the neural signal through time. In this case this is reflected by the capability to retain information learned during, for instance, the MRec40 session and re-use that information, in the form of a pre-trained model, for decoding MRec41. The goal is to reduce even more the need for time consuming data acquisition and re-training of the models, hence, in a long term perspective, lowering the daily effort for BMI patients and clinicians. 
    In this specific case, the spike sorting applied separately for each session is not helping, as it changes the number of channels and mixes them up preventing the straightforward re-use of a model. In this sense, an effort of applying this model to datasets from literature that are not spike-sorted per session is ongoing. Relying on public datasets will also provide an opportunity to validate this approach against a much wider prior literature.
\subsubsection*{Acknowledgements}
This work was supported by the European Union Horizon 2020 research and innovation program under Grant Agreement No. 965044
and by CINECA, through the Italian SuperComputing Resource Allocation - ISCRA-C grant.
We thank Stefan Schaffelhofer for data collection and Federico Fabiani for sharing the useful experience gained during his Master's Thesis work.

\subsubsection*{Ethics statement}
    A comprehensive discussion of the ethical aspect of BMIs is out of the scope of this paper: the B-Cratos project has a dedicated ethics advisory board and provides an ethics report to the European Commission yearly. With respect to animal experimentation, the data discussed here has been acquired long before the start of this work and they were subject to regulations and ethics assessment.

    On the other hand, ethics topics are particularly relevant when dealing with ML algorithms for BMI use: in this work, for instance, different hyperparameters configurations were needed by animal Z (female) and animal M (male). For a future clinical use, assuming that a single model architecture would fit all users may introduce unwanted biases and negatively affect the user experience of patients.

    Finally, several privacy and security aspects have been discussed in relation to this application: from anonymised data storage that can prevent the association between patients brain recordings and their identity, up to the possibility of adversarial actors affecting the prosthesis movement by exploiting weaknesses specifically related to the ML model (i.e., by injecting malicious training data) and the potential mitigation actions. These discussions are also reported in the ethics deliverable submitted to the EC and not yet publicly accessible.
%\subsubsection*{Acknowledgments}
%
%This work has been developed thanks to the C3S supercomputing facility and it has been partially supported by the OptiBike experiment in the H2020 projects Fortissimo2 (no. 680481)

\printbibliography

\end{document}